%% file: ms.tex
\documentclass[11pt]{article} 

\usepackage{graphicx}
\usepackage{tabularx}
\usepackage{booktabs}
\usepackage[T1]{fontenc}
\usepackage{lmodern}
\usepackage[flushleft]{threeparttable}
\usepackage[hidelinks]{hyperref} %
\hypersetup{hidelinks}
\usepackage{multirow}
\usepackage{longtable}
\usepackage[margin=3cm]{geometry} %
\usepackage{amsmath}
\usepackage{amssymb}
\usepackage{authblk}
\usepackage{caption}
\usepackage{flafter}

\title{A Robust Remote Photoplethysmography Method}
\author[1]{Alexey Protopopov}
\affil[1]{Joint Stock Research and Production Company Kryptonite \authorcr
E-mail: a.protopopov@kryptonite.ru}
\date{}
\begin{document}
    \captionsetup[figure]{labelformat={default},labelsep=period,name={Figure}}
    \captionsetup[table]{labelformat={default},labelsep=period,name={Table}}

    \maketitle

    \begin{abstract}
        Remote photoplethysmography (rPPG) is a method for measuring a subject’s heart rate remotely using a camera. Factors such as subject movement, ambient light level, makeup etc. complicate such measurements by distorting the observed pulse. Recent works on this topic have proposed a variety of approaches for accurately measuring heart rate in humans, however these methods were tested in ideal conditions, where the subject does not make significant movements and all measurements are taken at the same level of illumination. In more realistic conditions these methods suffer from decreased accuracy. The study proposes a more robust method that is less susceptible to distortions and has minimal hardware requirements. The proposed method uses a combination of mathematical transforms to calculate the subject’s heart rate. It performs best when used with a camera that has been modified by removing its infrared filter, although using an unmodified camera is also possible. The method was tested on 26 videos taken from 19 volunteers of varying gender and age. The obtained results were compared to reference data and the average mean absolute error was found to be at 1.95 beats per minute, which is noticeably better than the results from previous works. The remote photoplethysmography method proposed in the present article is more resistant to distortions than methods from previous publications and thus allows one to remotely and accurately measure the subject’s heart rate without imposing any significant limitations on the subject’s behavior.
    \end{abstract}

    \emph{Keywords}: Biomedical Monitoring, CIELAB, Photoplethysmography, Remote Heart Rate Monitoring, Remote Sensing.

    \input{full}

\end{document}

%% file: full.tex
\section{Introduction}\label{introduction}

Remote photoplethysmography (rPPG) uses cameras to measure the subtle changes in the person’s skin color and calculate the heart rate, which introduces a number of factors that complicate measurements, such as dependency on the intensity and position of ambient light sources, subject movement, etc. As a consequence, the majority of existing rPPG methods \cite{sinhal, salim} perform well only when the subject remains still and the lighting does not change during the measurement. For instance, the method presented in \cite{lernia} gives an average mean absolute error (MAE) of 5.50~BPM. Similarly, in \cite{haugg} the authors propose an algorithm that gives a MAE of 1.91~BPM for subjects at rest, and a MAE of 7.98~BPM for talking subjects. Finally, the algorithm proposed in \cite{artemyev} produces results with a MAE of 5.45~BPM for moving subjects. It is clear then, that subject movement presents a problem for rPPG methods. Thus, in this paper, we introduce a new method, which combines face feature detection, spectrogram calculation and iterative curve fitting to accurately determine the subject’s heart rate, significantly suppressing the effects of their movement, presence of makeup or changes in the surroundings.

\section{Methods}\label{methods}

To validate our findings, we produced 26 video recordings gathered from 8 male and 11 female volunteers of ages between 20 and 50 (informed consent was obtained from all participants). The volunteers were selected solely based on their willingness to participate in the experiment, they were not selected based on their ethnicity. Each of the recordings were approximately 5 minutes long, meaning that the total duration of video footage was just over 2 hours. We used a Logitech 720p web camera to produce these recordings, which was modified by removing the infrared (IR) filter. This was done because some types of makeup obscure the skin and as a result have a negative impact on measurement accuracy. By expanding the camera’s spectral range into the near-IR, we were able to bypass this problem. Reference heart rate measurements were performed with an Arduino-based heart rate monitor using reflective photoplethysmography. The relative amplitude of reflected light recorded by this monitor was subjected to a fast Fourier transform (FFT) to determine the true heart rate. These measurements were synchronized with the video recordings to provide an accurate reference.

\subsection{Color space}\label{meth_color_space}

Most works on the topic of rPPG utilize the regular RGB color space for extracting the heart rate from the video. However, using this color space makes the method very susceptible to subject movement, since it does not separate the hue of the surface from its luminosity. This introduces significant distortions as the signal levels in all channels change sharply every time the subject moves.

\begin{figure}[htbp]
    \centering
    \includegraphics[scale=0.7]{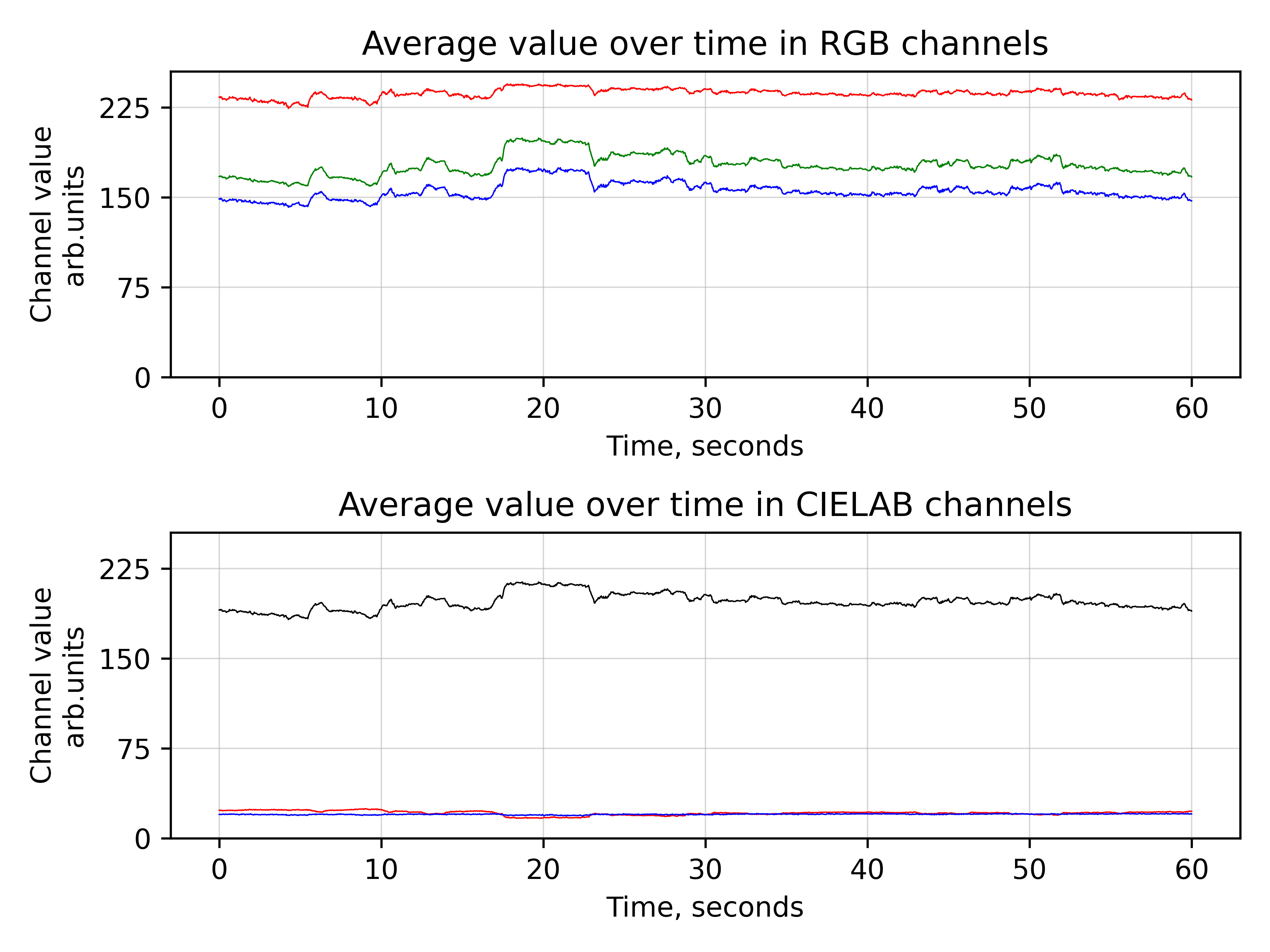}
    \caption{Comparison between RGB and CIELAB color spaces.}
    \label{fig:1}
\end{figure}

 In the present paper, we propose using the CIELAB color space, which separates luminosity into a separate channel, allowing us to operate directly with the hue. While this does not completely eliminate movement artifacts, it significantly reduces them. A comparison between the RGB and CIELAB values is shown on Figure~\ref{fig:1}. The top graph shows the values of red, green and blue channels averaged over an area of the frame. The line colors correspond with the colors of the channels. The bottom graph shows values of channels channels L (black), a* (red) and b* (blue), averaged over the same area of the frame.

The graphs in Figure~\ref{fig:1} were taken from a video fragment, where the subject moved around the frame. As can be easily seen, in case of the RGB color space, all three channels contain movement artifacts, while in case of the CIELAB color space, virtually all movement artifacts are localized in the luminosity channel, as expected. Thus, it is not used in our calculations. Furthermore, our experiments have shown that of the remaining two channels, only the a* channel contains the heartbeat signal, so the other one can be safely ignored.

\subsection{Face tracking}\label{meth_face_tracking}

\begin{figure}[htbp]
    \centering
    \includegraphics[scale=0.7]{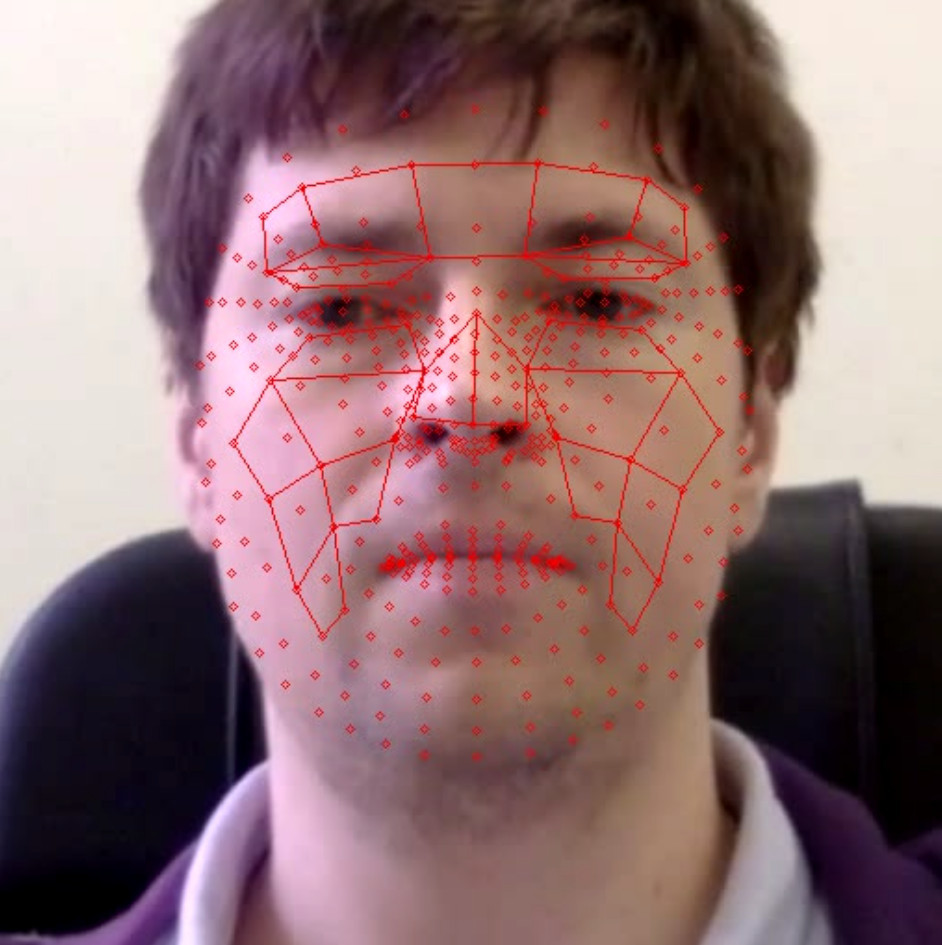}
    \caption{A subject’s face with landmarks drawn over it.}
    \label{fig:2}
\end{figure}

In this paper, we wanted to create a robust algorithm, that would not impose any constraints on the subject behavior (i.e. remaining still and looking at the camera). Therefore, a face tracking algorithm was implemented using the MediaPipe neural network. Conveniently, it not only tracks the position of the subject’s face, but also marks over 400 landmarks on it, as shown of Figure~\ref{fig:2}. These allow us to gather the signal from specific areas of the face (shown as a grid in the image), rather than from a box drawn around it, which helps further reduce motion artifacts.

These areas were selected based on how likely they were to be obstructed by facial hair or clothes, and on how likely they would be to face the camera. For each consecutive frame, the pixels belonging to the cells would have the value in their a* channel averaged according to the following formula:

\begin{equation}
    S_{n}(t)=\sqrt{\sum_{i,j=1}^{H,W}(s_{n,i,j})^{2}}
    \label{eq:1}
\end{equation}

where \emph{i} and \emph{j} are the indexes of the pixel within the frame along the horizontal and vertical axes, \emph{H} and \emph{W} are the frame’s height and width, \emph{n} is the number of the cell and \emph{s\textsubscript{n,i,j}} is the signal of the pixel’s a* channel. It is interesting to note, that the selection of the pixels that belong to the cells is the most computation-intensive step of the entire algorithm, accounting for approximately half of the running time.

\subsection{Signal processing}\label{meth_signal_processing}

\begin{figure}[htbp]
    \centering
    \includegraphics[scale=0.7]{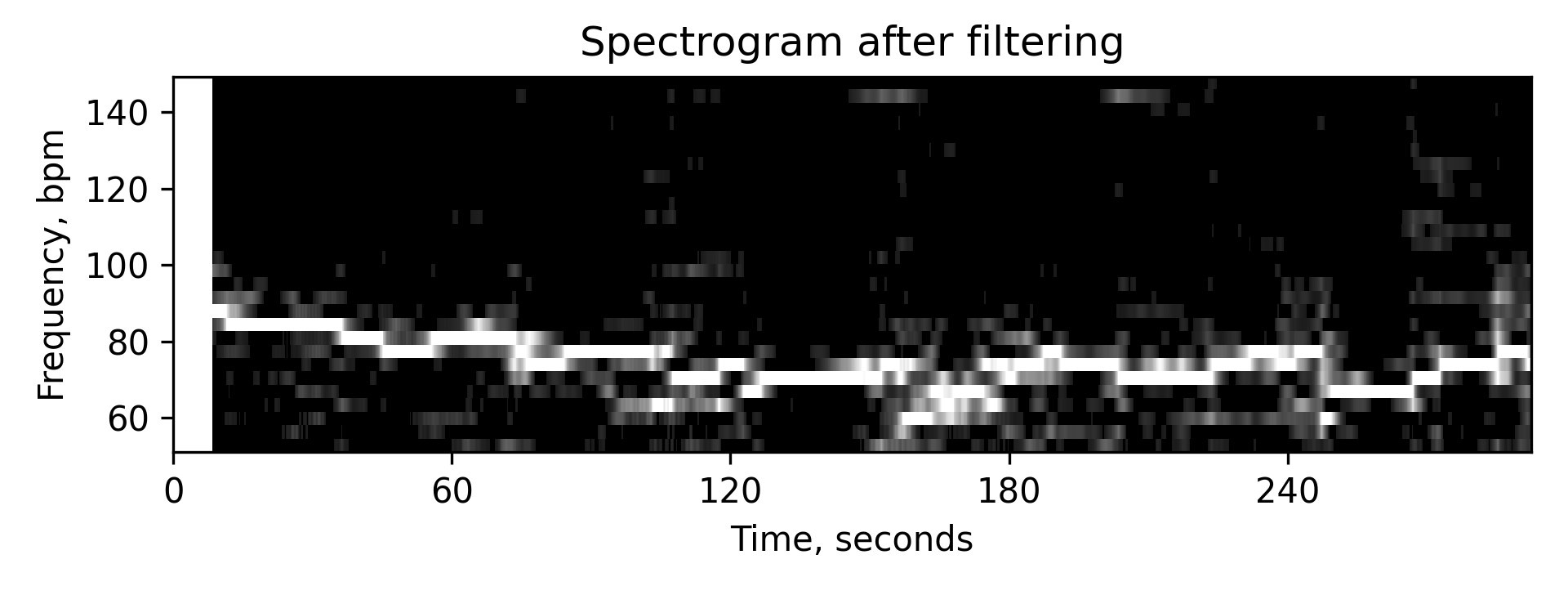}
    \caption{Spectrogram of the a* channel.}
    \label{fig:3}
\end{figure}

Following this, an FFT with a Tukey window is applied to the averaged signal, creating a spectrogram. This spectrogram is then normalized, so that the maximum signal in each time window (i.e. each column, if the time axis is horizontal) equals 1. Operating on the assumption that the harmonics corresponding to the heartbeat would have a sufficiently high amplitude compared to the rest of the harmonics, we eliminate most of the noise on the spectrogram by zeroing all values less than 0.1. The final spectrogram is shown on Figure~\ref{fig:3}. The frequency range is set to be between 50 and 150~bpm, as this is a reasonable range for a healthy human.

Having obtained the spectrogram, we need to select the harmonics corresponding to the heart rate. Simply choosing the maximum in a given time window is not an option, since in some cases subject movement may cause short-lived peaks in the spectrum, which may be interpreted incorrectly. Thus, we developed an iterative algorithm based on the PyTorch backpropagation function, which fits a polyline on top of the spectrogram. This polyline is then treated as the heart rate curve.

The algorithm begins by selecting the starting positions of the line’s vertexes. The X-coordinates of the vertexes are fixed and placed equidistant along the horizontal axis. The starting Y-coordinates are set to be the same for all vertexes and are calculated using the following formula:

\begin{equation}
    y_{start}=\sum_{k=1}^{K}\left(y_{k}\sum_{l=1}^{L}P_{k,l}  \right)/\sum_{k,l=1}^{K,L}P_{k,l}
    \label{eq:2}
\end{equation}

where \emph{k} is the index along the frequency axis, ranging from 1 to \emph{K}, \emph{l} is the index along the time axis, ranging from 1 to \emph{L}, \emph{P\textsubscript{k,l}} is the power of the corresponding harmonic, \emph{y\textsubscript{k}} is the coordinate of the corresponding harmonic along the frequency axis.

After the polyline’s initial position is set, the algorithm uses an ADAM optimizer to minimize the loss function by adjusting the values of the polyline’s Y-coordinates. The loss function is given by the following formulas:

\begin{equation}
    L=\alpha\cdot L_{p}-\frac{L_{n}}{\alpha}
    \label{eq:3}
\end{equation}
\begin{equation}
    L_{P}=\sum_{m=1}^{M-1}\left| \Delta y_{m} \right|^{\beta}
    \label{eq:4}
\end{equation}
\begin{equation}
    \Delta y_{m}=y_{m+1}-y_{m}
    \label{eq:5}
\end{equation}
\begin{equation}
    L_{n}=\sum_{m=1}^{M}\left[ \sum_{k=1}^{K}\left( w_{k,m}\cdot \sum_{l=x_{m}-p}^{x_{m}+p}P_{k,l} \right) \right]^{2}
    \label{eq:6}
\end{equation}
\begin{equation}
    w_{k,m}=\frac{1}{1+\frac{\left( k-y_{m} \right)^{2}}{K\cdot r^{2}}}
    \label{eq:7}
\end{equation}

\begin{figure}[htbp]
    \centering
    \includegraphics[scale=0.7]{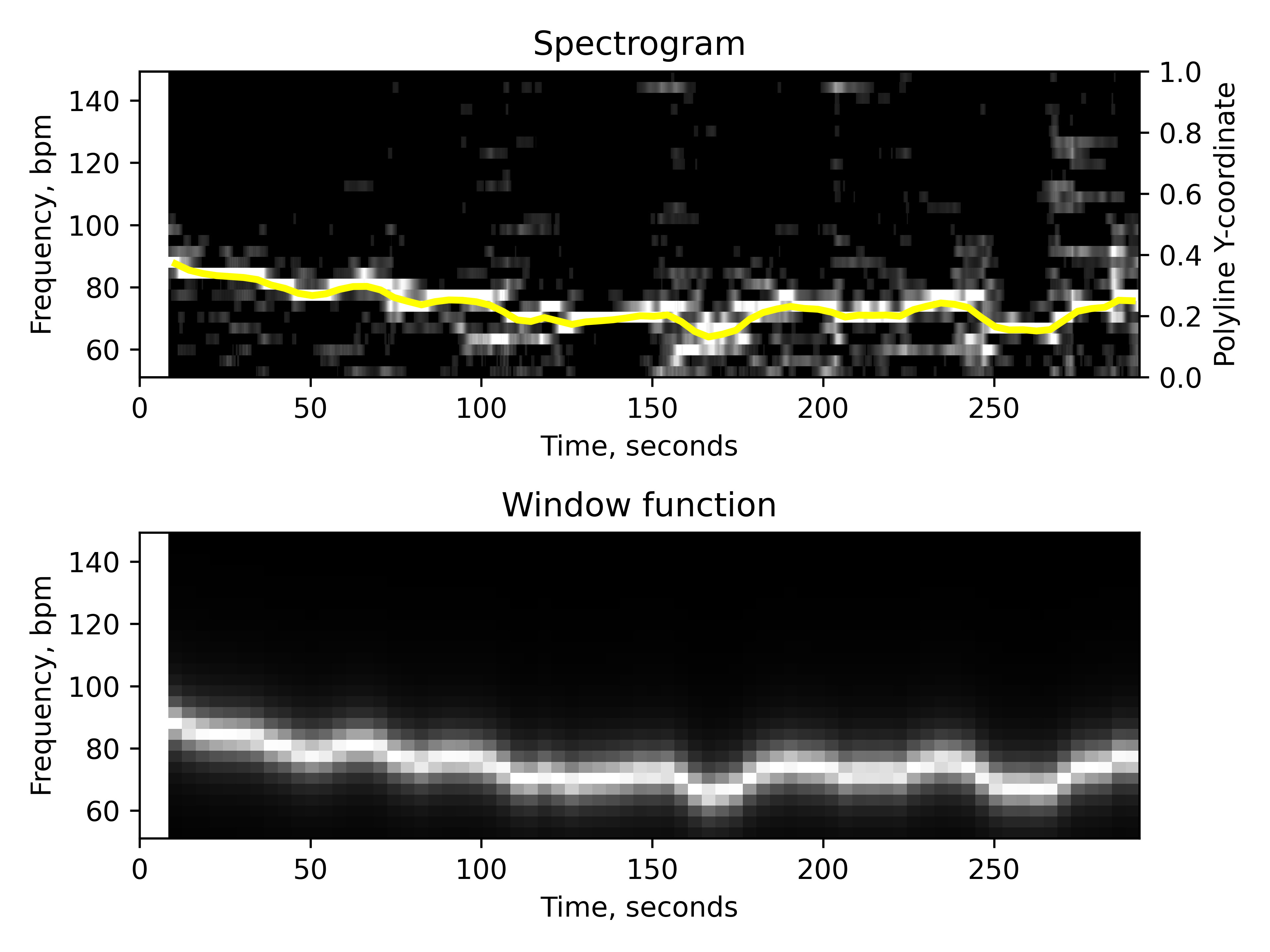}
    \caption{The spectrogram together with the map of the window function.}
    \label{fig:4}
\end{figure}

Here \emph{m} and \emph{M} are the index and total number of polyline vertexes, \emph{p} is an integer value equal to half the number of windows between each polyline vertex, \emph{x\textsubscript{m}} and \emph{y\textsubscript{m}} are the coordinates of polyline vertexes along the X and Y-axes, ranging between [0;~\emph{L}] and [0;~\emph{K}] respectively, and \emph{$\alpha$}, \emph{$\beta$} and \emph{r} are tuning values. Typically, it takes less than 100 iterations for the polyline to stabilize. One such result is shown on Figure~\ref{fig:4}, together with the map of \emph{w\textsubscript{k,m}}.

In essence, the loss function encourages the polyline to place itself over the harmonics with higher amplitude, while at the same time discouraging it from becoming longer.

\begin{table}
    \caption{Experimental results.}
    \label{tab:1}
    \centering
    \begin{tabular}{||c|c|c||c|c|c||}
        \hline
        Experiment & Subject & MAE, bpm & Experiment & Subject & MAE, bpm \\
        \hline
        1 & 1 & 0.63 & 14 & 7 & 1.29 \\
        \hline
        2 & 1 & 1.57 & 15 & 8 & 2.86 \\
        \hline
        3 & 1 & 0.32 & 16 & 9 & 3.79 \\
        \hline
        4 & 2 & 7.80 & 17 & 10 & 1.20 \\
        \hline
        5 & 2 & 7.41 & 18 & 11 & 1.26 \\
        \hline
        6 & 2 & 5.13 & 19 & 12 & 0.45 \\
        \hline
        7 & 3 & 0.74 & 20 & 13 & 1.12 \\
        \hline
        8 & 3 & 0.46 & 21 & 14 & 1.40 \\
        \hline
        9 & 4 & 0.98 & 22 & 15 & 1.78 \\
        \hline
        10 & 4 & 3.26 & 23 & 16 & 1.22 \\
        \hline
        11 & 4 & 1.90 & 24 & 17 & 0.39 \\
        \hline
        12 & 5 & 0.63 & 25 & 18 & 0.73 \\
        \hline
        13 & 6 & 0.47 & 26 & 19 & 1.77 \\
        \hline
    \end{tabular}
\end{table}

\section{Results and discussion}\label{results}

Now that we have acquired the heart rate values from the camera, we may compare them to the reference heart rate. The results of this comparison are shown in Table~\ref{tab:1}. We use the MAE between the experimental value and the reference to quantify the quality of our measurements. 

\begin{figure}[htbp]
    \centering
    \includegraphics[scale=0.7]{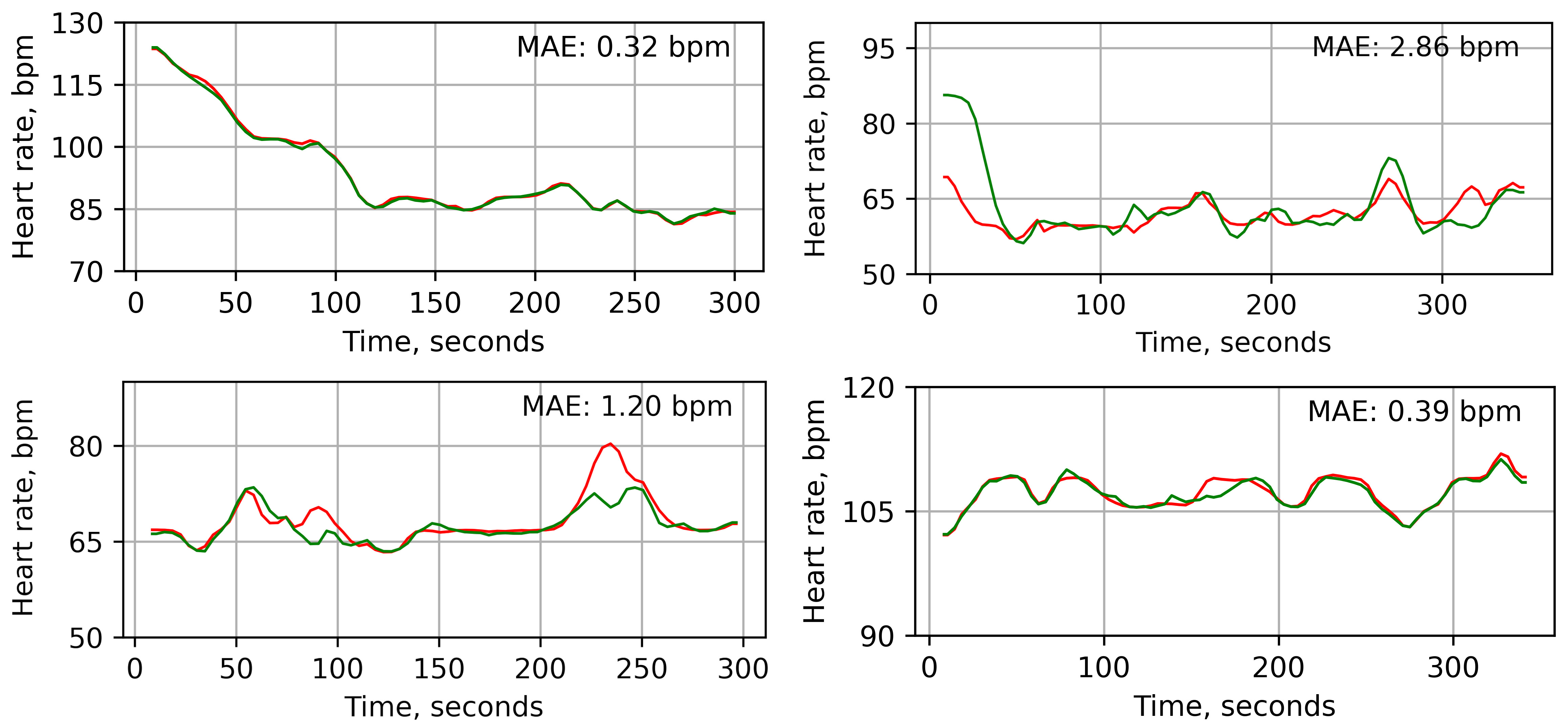}
    \caption{Heart rate comparison plots.}
    \label{fig:5}
\end{figure}

Since all our videos are of similar length, we may calculate the average MAE by taking the mean of these values, which is 1.95~BPM. For comparison, in similar experimental conditions \cite{lernia}, \cite{haugg} and \cite{artemyev} give the MAE of 5.50, 7.98 and 5.45~BPM respectively, which means that the method proposed in the present article offers greater resistance to distortions caused by movement, makeup etc. Some of the typical heart rate curves are shown in Figure~\ref{fig:5}. The correlation between the heart rate measured from the video (green line) and the reference heart rate (red line) is clearly visible. MAE is shown in the top right corner of each plot.

As can be seen in Table~\ref{tab:1}, the heart rate produced by the proposed method closely matches the reference despite the fact that these results were acquired without imposing any constraints on the subjects: they were not limited in terms of facial expressions and were free to look around. The only restriction was that the subject should not leave the camera’s field of view or look away for an extended period of time. At the same time, certain discrepancies between the heart rates were observed. Most of these were situated around the edges of the video and may be considered to be edge artifacts, since they would not have occurred if the video had been extended: the spectrum peaks corresponding to the heart rate would have “dragged” the line to a correct value.

It should be noted that our method has low hardware requirements. All measurements were performed using a low-cost web camera with minimal modification. However, this created certain limitations: since the camera’s frame rate is only 30~fps, and to get a reasonable resolution along the frequency axis we needed to compute the FFT over 512~windows, this meant that each window was just over 17~seconds wide. And while the window stride was 4~frames, or less than 0.2~seconds, the algorithm would not be able to pick up short-term spikes in the heart rate. Naturally, with a more advanced camera capable of higher frame rate, this limitation would be less severe.

\section{Conclusion}\label{conclusion}

To summarize the present paper, the proposed method has the following advantages:

\begin{enumerate}
    \item Robustness. The method displays greater resistance to subject movement and makeup when compared to existing works on the same subject;
    \item Low hardware requirements. Most existing web cameras are suitable for the algorithm, provided the IR filter is removed. It should be noted that this method does work even with unmodified cameras, however in this case the accuracy of heart rate measurement suffers, especially if the subject has applied a significant amount of makeup to the face;,
    \item Low computation complexity. In our tests conducted on an Intel~i5-10210U~CPU, the total running time of the algorithm was approximately equal to 1.5~times the duration of the video recording loaded into it. It should be noted, that nearly half of that time was spent on determining which pixels of the frame correspond to the selected areas of the face.
\end{enumerate}

At the same time, the method has the following drawbacks:

\begin{enumerate}
    \item The algorithm does not register the heart rate in real time;
    \item The resolution along the time axis is limited.
\end{enumerate}

The first limitation is caused by the final step in the algorithm. Since it uses an iterative approach to fit a polyline over the spectrogram, it requires the entire length of the spectrogram along the time axis. One of the topics of future works on this subject could be the improvement of this step to work on spectrograms that are being extended in real time. The second limitation is caused by the finite framerate of the camera used to film the subject. One way to overcome this would be to use a camera with a higher framerate. Another, potentially cheaper option, would be to use several cameras working in parallel to increase the effective framerate of the system.

\section{Acknowledgements}\label{acknowledgements}

I would would like to thank my supervisor Vasiliy Dolmatov for his guidance and support, which was a great help for me in my research. I would also like to thank Jurinflot International Law Firm for providing many volunteers for testing the method outlined in the present article.

\clearpage

\bibliographystyle{ieeetr}
\bibliography{ms}